\documentclass[conference]{IEEEtran}
\IEEEoverridecommandlockouts
\usepackage{cite}
\usepackage{amsmath,amssymb,amsfonts}
\usepackage{graphicx}
\usepackage{textcomp}
\usepackage{xcolor}

\usepackage{algpseudocode}
\usepackage{algorithm}
\algrenewcommand\algorithmicindent{0.7em}         %  https://tex.stackexchange.com/a/47703
\algnewcommand{\LineComment}[1]{\State \(\triangleright\) #1}

\usepackage{booktabs}
\usepackage{subfigure}

\usepackage{booktabs}
\usepackage{graphicx}

\usepackage{caption}
\def\BibTeX{{\rm B\kern-.05em{\sc i\kern-.025em b}\kern-.08em
    T\kern-.1667em\lower.7ex\hbox{E}\kern-.125emX}}
\begin{document}

\title{Unsupervised Replay Strategies for Continual Learning with Limited Data}

\author{\IEEEauthorblockN{Anthony Bazhenov$^*$}
\IEEEauthorblockA{
%\textit{dept. name of organization (of Aff.)} \\
\textit{Del Norte High School}\\
San Diego, CA, USA \\
anthonyb11c234@gmail.com}

\and
\IEEEauthorblockN{Pahan Dewasurendra$^*$}
\IEEEauthorblockA{
%\textit{dept. name of organization (of Aff.)} \\
\textit{Del Norte High School}\\
San Diego, CA, USA \\
pahan.dewa@gmail.com}

\and
\IEEEauthorblockN{Giri P. Krishnan}
\IEEEauthorblockA{\textit{Department of Medicine} \\
\textit{University of California, San Diego}\\
La Jolla, CA, USA \\
giri.prashanth@gmail.com}

\and
\IEEEauthorblockN{Jean Erik Delanois}
\IEEEauthorblockA{\textit{Department of Computer Science \& Engineering} \\
\textit{University of California, San Diego}\\
La Jolla, CA, USA \\
jdelanois@ucsd.edu}

}
% \author{\IEEEauthorblockN{Anonymous Authors}}

\maketitle
\begingroup\renewcommand\thefootnote{\textsection}
\footnotetext{ $^*$Equal contribution}

\footnotetext{DOI: 10.1109/IJCNN60899.2024.10650116}

\footnotetext{© 2024 IEEE.  Personal use of this material is permitted.  Permission from IEEE must be obtained for all other uses, in any current or future media, including reprinting/republishing this material for advertising or promotional purposes, creating new collective works, for resale or redistribution to servers or lists, or reuse of any copyrighted component of this work in other works.}

\endgroup

\begin{abstract}
Artificial neural networks (ANNs) show limited performance with scarce or imbalanced training data and face challenges with continuous learning, such as forgetting previously learned data after new tasks training. In contrast, the human brain can learn continuously and from just a few examples. This research explores the impact of 'sleep'—an unsupervised phase incorporating stochastic activation with local Hebbian learning rules—on ANNs trained incrementally with limited and imbalanced datasets, specifically MNIST and Fashion MNIST. We discovered that introducing a sleep phase significantly enhanced accuracy in models trained with limited data.  When a few tasks were trained sequentially, sleep replay not only rescued previously learned information that had been catastrophically forgetting following new task training but often enhanced performance in prior tasks, especially those trained with limited data. This study highlights the multifaceted role of sleep replay in augmenting learning efficiency and facilitating continual learning in ANNs.
\end{abstract}


\section{Introduction}

Artificial neural networks have demonstrated the ability to excel beyond human levels in various domains. However, they encounter difficulties when training data are low, imbalanced or presented in sequentially where they tend to prioritize new tasks at the expense of previous tasks, a phenomenon called catastrophic forgetting \cite{RN4279, hayes2021replay}. In contrast, humans and animals possess an extraordinary capacity to learn continuously and just from few examples, effortlessly assimilating new information into their existing knowledge base.

%Deep learning methods have shown considerable performance when training datasets are large, however, existing techniques generally fail in low training data conditions. Additionally, training datasets are often imbalanced, often due to some categories naturally occurring less frequently than others, resulting in reduced accuracy for ANNs. 

Different methods have been proposed to overcome these limitations. For low training data scenarios, these include data augmentation \cite{shorten2019survey}, pre-training on other datasets \cite{zhuang2020comprehensive} or alternative architectures such as neural tangent kernel \cite{arora2019harnessing}. However, these approaches do not address the fundamental question of how to make overparameterized deep learning networks learn to generalize from small datasets without overfitting. 

Existing approaches to prevent catastrophic forgetting generally fall under two categories: rehearsal and regularization methods \cite{kemker2018measuring}. Rehearsal methods combine previously learned data, either stored or generated, with novel inputs in the next training to avoid forgetting \cite{hayes2019memory, robins1995catastrophic, van2020brain,shin2017continual, buzzega2020dark, buzzega2021rethinking}. Regularization approaches \cite{li2017learning,kirkpatrick2017overcoming} aim to modify plasticity rules by incorporating additional constraints on gradient descent such that important weights from previously trained tasks are maintained. All these methods have significant limitations (reviewed in \cite{RN4279, hayes2021replay}). Importantly, methods to prevent catastrophic forgetting are not always compatible with methods aimed at improving low data performance.

The human brain demonstrates the ability to learn continuously and quickly from just a few examples. It has been suggested that memory replay during biological sleep can strengthen memories learned during wakefulness
\cite{stickgold2005sleep, lewis2011overlapping}. During sleep, neurons are spontaneously active without external input and generate complex patterns of synchronized activity across brain regions \cite{steriade1993thalamocortical, krishnan2016cellular}. 
Two critical components which are believed to underlie memory consolidation during sleep are spontaneous replay of memory traces and local unsupervised synaptic plasticity that  restricts synaptic changes to relevant memories only \cite{wilson1994reactivation, stickgold2005sleep, wei2016synaptic}.  
During sleep, replay of recently learned memories along with relevant old memories \cite{lewis2018memory, hennevin1995processing, rasch2013sleep, mednick2011opportunistic, paller2004memory, oudiette2013role}
enable the network to form stable long-term memory representations \cite{rasch2013sleep} and reduce competition between orthogonal memory representations to enable coexistence of competing memories within overlapping populations of neurons 
\cite{gonzalez2020can}. 

The idea of replay has been explored in machine learning to enable continual learning (reviewed in \cite{RN4279, hayes2021replay}). However, spontaneous unsupervised replay found in the biological brain is significantly different compared to explicit replay of past inputs implemented in machine learning rehearsal methods. Unlike the standard global optimization methods used in machine learning (such as backpropagation), local plasticity allows synaptic changes to affect only relevant memories.  Research from neuroscience \cite{wei2016synaptic, gonzalez2020can, golden2020sleep,Tadros_NC2022} suggest that applying sleep replay principles to ANNs may enhance memory representations and, consequently, improve the performance of machine learning models trained continuously on a limited or unbalanced datasets.

In this new study, we apply the previously proposed sleep replay consolidation (SRC) method \cite{Tadros_NC2022} to scenarios when the model is trained sequentially on limited data. We found that sleep-like replay can both improve the performance of models trained with limited data as well as rescue previously trained task performance that was damaged by new training.

\section{Methods}

\subsection{Neural Network}\label{sec:ANN}

We used a fully connected feed-forward neural network with 2 hidden layers and ReLU nonlinearities consisting of 1200 nodes each, followed by a soft-max classification layer with 10 output neurons. 
%While the network was operating in an Artificial Neural Network (ANN) regime, hidden layers leveraged ReLU nonlinearities.
The model was trained using hidden layer dropout and a binary cross entropy loss with weights modified by a standard stochastic gradient descent optimizer. Each neuron in the network operated without a bias, which aided in the conversion to a spiking neural network during the sleep stage \cite{diehl2015fast}. 

%When the ANN was converted to a Spiking Neural Network (SNN) for sleep, all activation functions were replaced with the Heaviside thresholding function to enable spiking behavior while the weight matrices remained fixed, thereby preserving the structure developed by the ANN during the original training before sleep. More details about the sleep phase are provided in Section \ref{sec:SRC}. A summary of the network parameters are shown in Table \ref{table:nntrain}. 

We used 5 epochs of training in the main analysis and we verified results using 10 and 50 epochs of training. 
%These long training duration experiments allowed us to observe that while increasing the amount of training improved baseline performance in data restricted scenarios, sleep was still capable of further improving accuracy. 
The fine-tuning stage following sleep consisted of a single epoch of training with a learning rate of 0.02 with all other hyperparameters remaining the same. The same network architecture and training parameters were used in all simulations.

% \usepackage{booktabs}
% \usepackage{graphicx}
\begin{table}[]
\resizebox{\columnwidth}{!}{%
\begin{tabular}{@{}ccccc@{}}
\toprule
\multicolumn{2}{c}{Model Details}                 &  &  &  \\ \midrule
Arch. Size                           & 784, 1200, 1200, 10 &  &  &  \\
Learning Rate (Baseline / Fine-tune) & 0.06 / 0.02         &  &  &  \\
Epochs (Baseline / Fine-tune)        & 5 (10, 50) / 1      &  &  &  \\
Dropout                              & 0.25                &  &  &  \\ \bottomrule
\end{tabular}%
}
\caption{Neural network model and training parameters}
\label{table:nntrain}
\end{table}

\subsection{Sleep Replay Consolidation (SRC) algorithm}\label{sec:SRC}

The fully-connected ANN with two hidden layers described above was first trained 
%on a randomly selected subset of MNIST or Fashion MNIST (FMNIST) datasets 
using backpropagation. Subsequently, the sleep replay consolidation (SRC) algorithm was implemented using this trained model, as previously described in \cite{Tadros_NC2022}. 

The intuition behind SRC is that a period of off-line, noisy activity may reactivate network nodes that represent  tasks trained in awake. If network reactivation is combined with unsupervised learning, SRC can then strengthen necessary and weaken unnecessary pathways in the network. Even if a model is under-trained, since there is information regarding the task encoded in the synaptic weight matrix, SRC can augment and enhance this previously existing structure through replay.

To implement SRC, the ANN trained by limited data in the first training phase was mapped to a spiking neural network (SNN) with the same architecture. During this mapping of  ANN to SNN, the network's activation function is replaced by a Heaviside function and weights are scaled by the layer-wise activation maximum observed on the training dataset, as suggested by \cite{diehl2015fast}, to ensure the network maintains reasonable firing activity.

The SRC phase starts with a forward pass where Poisson distributed spike trains reflective of input pixel averages are generated and fed to the input layer in order to propagate activity (spiking behavior) across the network. Following the forward pass, a backward pass is executed to update synaptic weights. To modify network connectivity during sleep we use an unsupervised, simplified Hebbian type learning rule which is implemented as follows: a weight is increased between two nodes when both pre- and post-synaptic nodes are activated (i.e., input exceeds the Heaviside activation function threshold), a weight is decreased between two nodes when the post-synaptic node is activated but the pre-synaptic node is not (in this case, presumably another pre-synaptic node is responsible for activity in the post-synaptic node). After running multiple steps of this unsupervised learning during sleep, the final weights are rescaled again (simply by removing the original scaling factor), the Heaviside-type activation function is replaced by ReLU, and testing or further supervised training on new data is performed. This all is implemented by a simple SRC function call after the initial baseline training. The hyperparameters used during SRC are listed in Table \ref{tab:hparams_SRC} and were determined using a genetic algorithm aimed at maximizing performance on the validation set (results then generalized to the test set).

As we mentioned earlier, to ensure network activity during SRC, the input layer of the network was activated with noisy binary (0/1) inputs. In each input vector (i.e., for each forward SRC pass), the probability of assigning a value of 1 (bright or spiking) to a given element (input pixel) is taken from a Poisson distribution with mean rate calculated as a mean intensity of that input element across all the inputs observed during all of the preceding training sessions. Thus, e.g., a pixel that was typically bright in all training inputs would be assigned a value of 1 more often than a pixel with lower mean intensity. 

%In \cite{Tadros_NC2022}, SRC was applied in the scenario of sequential task training after each new task to avoid catastrophic forgetting. Here, we applied SRC once after training the ANN with limited data.

\subsection{Datasets}

%It is well understood that deep learning models require significant amounts of data to achieve top-tier performance and that their performance degrades heavily when data are limited. 
To evaluate the effects of sleep on under-trained and under-performing Artificial Neural Networks (ANNs) we leveraged two datasets: MNIST \cite{lecun1998gradient} and Fashion MNIST (FMNIST) \cite{xiao2017fashion}. The MNIST dataset consists of handwritten numbers (0-9) each belonging to its own class while FMNIST consists of 10 classes of Zalando's article images. Together the MNIST and FMNIST datasets are some of the most widely used datasets in machine learning making them good candidates to initially investigate sleep's ability to improve performance in limited data and imbalanced data contexts. Each dataset consists of 60,000 training images and 10,000 testing images. 

\subsection{Single taks training with limited data}

To test the effect of sleep on networks trained with limited data, we trained the neural network on a randomly selected subset (0.1\% - 100\%) of the full datasets MNIST and FMNIST. Subsequently we applied a sleep replay and fine-tuning stage to see how performance was impacted. The sleep replay stage followed the algorithm explained in Section \ref{sec:SRC}.

During our experiment implementation, we found that the number of images in each class can vary significantly when a small fraction of data from MNIST or FMNIST is extracted randomly.  This can inadvertently cause preferential performance on over-represented classes and diminished performance on under-represented classes. Therefore, in our experiments we first ensured that each class contained the same exact number of images when a portion of data from MNIST or FMNIST was extracted.

% \usepackage{graphicx}
\begin{table}
\resizebox{\columnwidth}{!}{%
\begin{tabular}{c|cc}
\toprule
                         & MNIST                   & FMNIST                    \\ \hline
Time Steps               & 365                     & 267                       \\
Dt                       & 0.001                   & 0.001                     \\
Max Firing Rate (Input Generation)         & 211.9                   & 465.5                     \\
Alpha Scale (Augments layer-wise scales)      & 15.5                    & 23.7                      \\
% Alpha             & {[}0.57, 0.89, 38.77{]} & {[}0.40, 1.07, 28.22{]}   \\
Beta              & {[}24.3, 5.08, 18.42{]} & {[}22.60, 14.93, 23.05{]} \\
Decay                    & 0.97                    & 0.95                      \\
Synaptic Increase & $7.49 * 10^{-4}$        & $4.95 * 10^{-4}$          \\
Synaptic Decrease & $-1.87 * 10^{-4}$       & $-2.72 * 10^{-4}$     \\ \bottomrule
\end{tabular}%
}
\caption{Hyperparameters for the SRC algorithm}
\label{tab:hparams_SRC}
\label{tab:my-table}
\end{table}

% ALGORITHM
\begin{algorithm*}[!h]
% \caption{\textbf{: Sleep Replay Consolidation}}\label{alg:sleep}
\begin{algorithmic}[1]
% \Procedure{ANNtoSleepANN}{$nn$}
%   \State Change ReLU activation in CNN to Heaviside function in SleepANN and determine layer-wide specific threshold
%   \State Apply weight normalization and return scale, threshold for each layer
%   \Return $SleepANN, scales, threshold$
% \EndProcedure
% \Procedure{SleepANNtoANN}{$nn$}
%   \State Directly map the new, unscaled weights from Heaviside-network (SleepANN) to ReLU network (ANN)
%   \Return $ann$
% \EndProcedure
\footnotesize{
\Procedure{Sleep}{$nn, I, scales, thresholds$} \Comment{$I$ is input}
  \State Initialize $v$ (voltage) $=0$ vectors for all neurons
  \For{$t \gets 1 \textrm{ to } Ts$} \Comment{$Ts$ - Number of time steps during sleep}
    \State $ \mathbf{S} \gets 0s$
    \State $\mathbf{S}(1) \gets \textrm{Convert input $I$ to Poisson-distributed spiking activity}$
    \State $\mathbf{S}$ = ForwardPass($\mathbf{S}, v, \mathbf{W}, scales, thresholds$)
    \State $\mathbf{W}$ = BackwardPass($\mathbf{S},\mathbf{W}$)
  \EndFor
%  \Return $nn$ \Comment{Do we need this line?}
\EndProcedure
\Procedure{ForwardPass}{$S, v, W, scales, threshold$} 
  \For{$l \gets 2 \textrm{ to } n$} \Comment{n - number of layers}
  \State $\alpha \gets scales(l-1)$
  \State $\beta \gets threshold(l)$
  \State $ v(l) \gets \lambda v(l) + (\alpha * \mathbf{W}(l,l-1) * \mathbf{S}(l-1))$   \Comment{W(l,l-1) - weights / $\lambda$ - decay rate}
  % \State $ \mathbf{S}(l) \gets v(l)  > \beta $
  % \State $ \mathbf{S}(l)(v(l)  > \beta ) \gets 1 $
  \State $ \mathbf{S}(l)_i \gets 1$  $\forall$ $i$ where $v(l)_i  > \beta $
  \Comment{Propagate spikes}
  % \State $v(l)(v(l) > \beta) \gets 0$ \Comment{Reset spiking voltages}or \\
  \State $v(l)_i \gets 0$ $\forall$ $i$ where $v(l)_i  > \beta $ \Comment{Reset spiking voltages} \\
  \EndFor \\
  \Return $S$
\EndProcedure
\Procedure{Backward Pass}{$S,W$} 
  \For{$l \gets 2 \textrm{ to } n$}        \Comment{$n$ - number of layers}
  \State $ \mathbf{W}(l,l-1)_{i,j} \gets \scriptsize{\begin{cases} \mathbf{W}(l,l-1)_{i,j} + inc & \forall$  i,j $\mbox{where }  \mathbf{S}(l)_j = 1 \, $ $\& $ $\, \mathbf{S}(l-1)_i = 1 \\ \mathbf{W}(l,l-1)_{i,j} - dec & \forall$  i,j $\mbox{where } \mathbf{S}(l)_j = 1 \, $ $\& $ $\,  \mathbf{S}(l-1)_i = 0 \\ \mathbf{W}(l,l-1)_{i,j} &$   $\mbox{else } \end{cases}}$ \Comment{STDP}
  %\State Apply STDP to update weights (If neuron in layer $l$ spikes, increase weight from layer $l-1$ if neuron in $l-1$ spikes and decrease weight if neuron in $l-1$ does not spike in same time step).
  \EndFor \\
  \Return $W$
\EndProcedure
}
\end{algorithmic}

\caption{\textbf{Sleep Replay Consolidation}\cite{Tadros_NC2022}}\label{alg:sleep}
\label{alg:SrcAlgo}
\end{algorithm*}

\begin{figure}[t!]
\centering
\includegraphics[width=0.9\columnwidth]{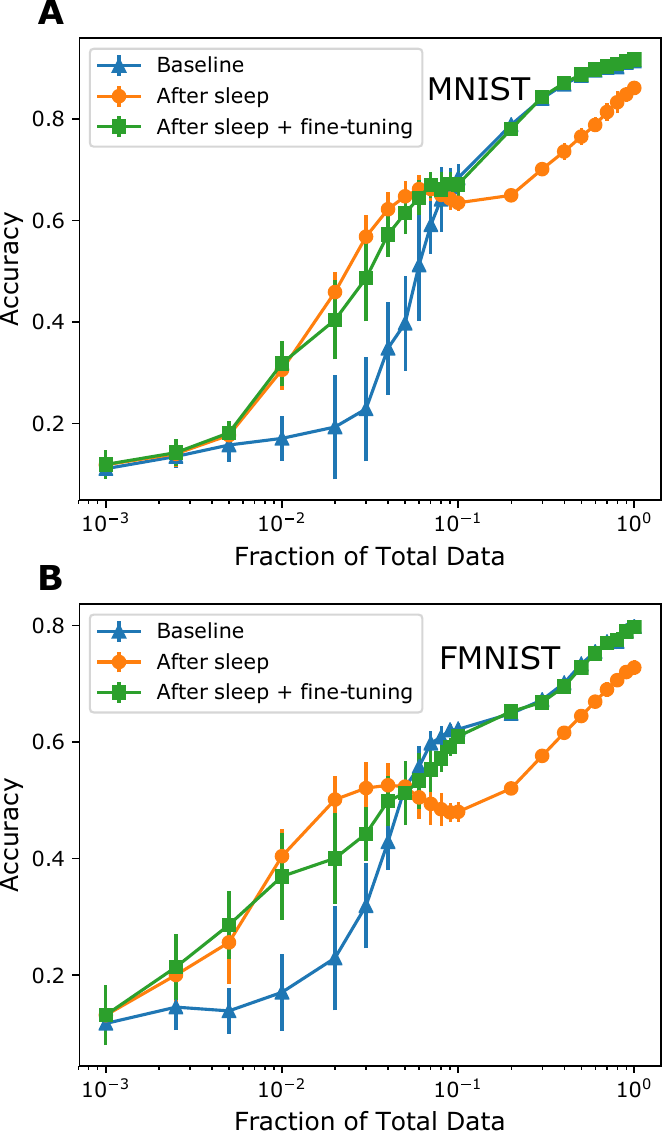}
\caption{Accuracy on MNIST \textbf{(A)} and FMNIST \textbf{(B)} with mean (lines) and standard deviation (error bars) across 10 trials. X-axis - log of the relative amount of data used for training (e.g., 0.01=1\% of data). Blue - baseline (after ANN training); Orange - baseline + sleep; Green - baseline + sleep + fine-tuning. Note significant gain in accuracy after the sleep phase on low data. The sleep phase reduced performance on high data but was largely recovered by fine-tuning.
}
\label{fig:fig1}
\end{figure}

\begin{figure}[!htbp]
\centering
\includegraphics{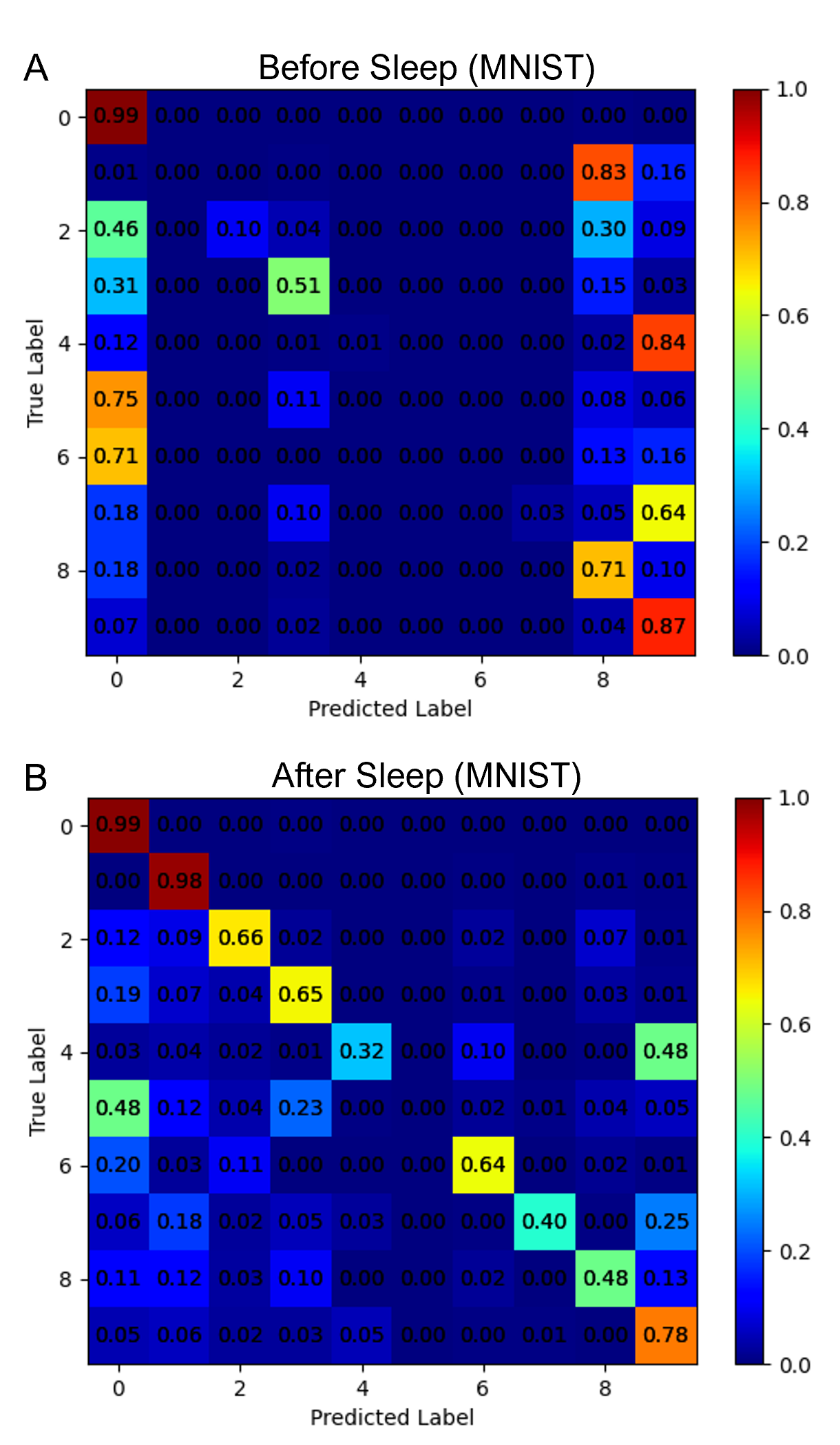} 
\caption{Confusion matrices before and after sleep for MNIST dataset. A 3\%  subset of the overall MNIST dataset was used in training. The value in each cell indicates the fraction of images of a given true label that were classified as a given predicted label by the model. \textbf{(A)} - before SRC, \textbf{(B)} - after SRC.}
\label{fig:confusion_matrix_MNIST}
\end{figure}

\begin{figure}[!htbp]
\centering
\includegraphics{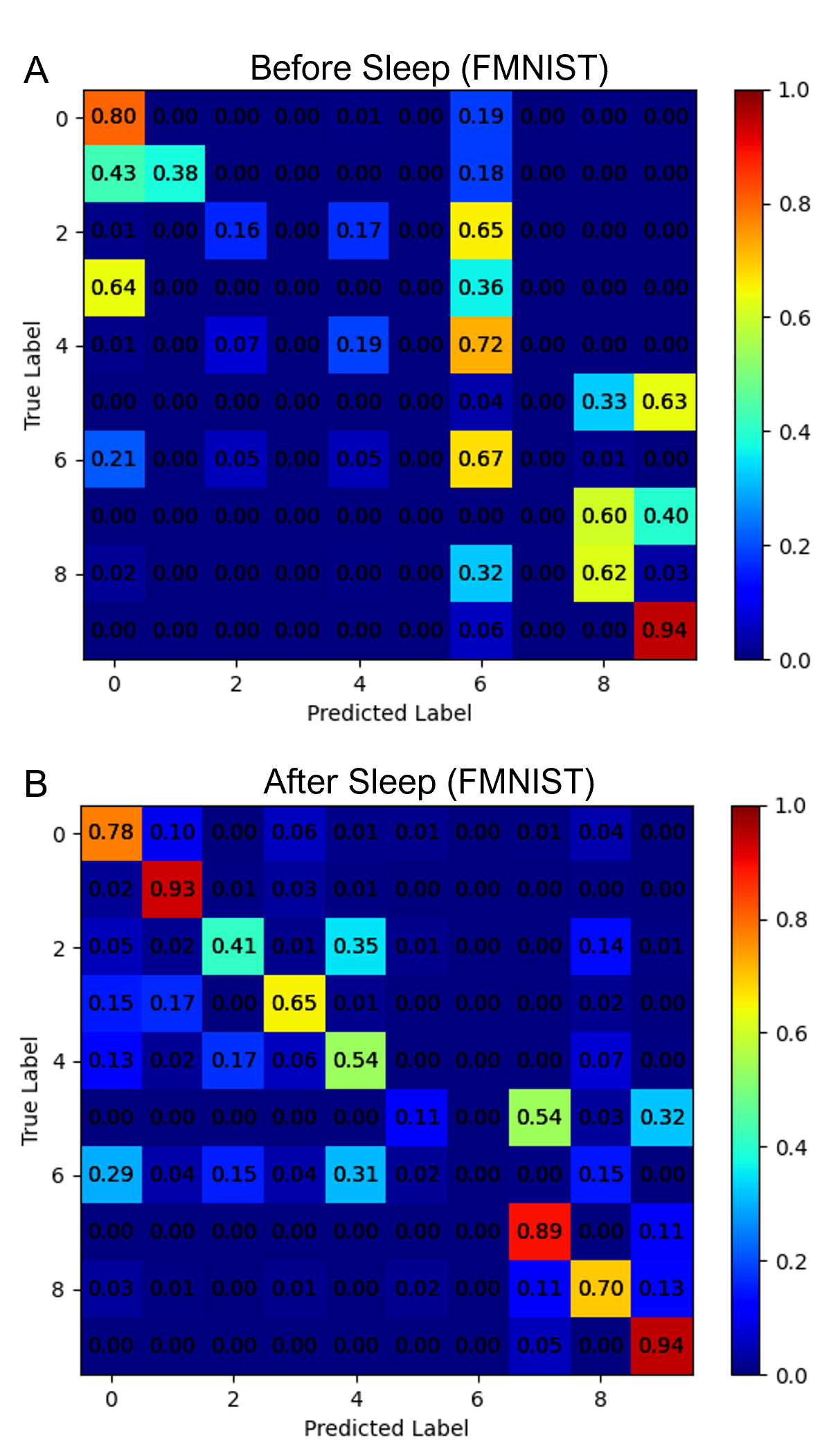} 
\caption{Confusion matrices before and after sleep for FMNIST dataset. A 3\%  subset of the overall FMNIST dataset was used in training. The value in each cell indicates the fraction of images of a given true label that were classified as a given predicted label by the model. \textbf{(A)} - before SRC, \textbf{(B)} - after SRC.}
\label{fig:confusion_matrix_FMNIST}
\end{figure}

\begin{figure}[t]
\centering
\includegraphics[width=1.0\columnwidth]{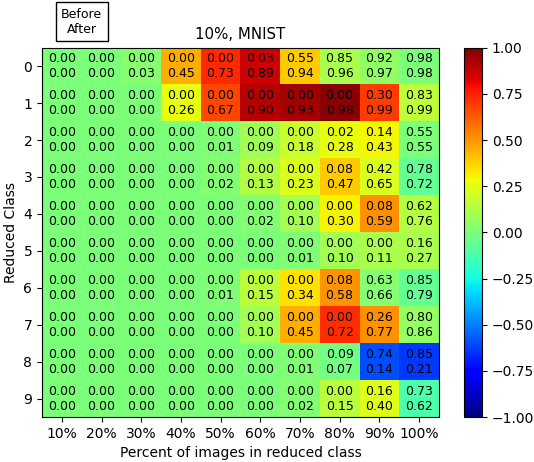} 
\caption{Imbalanced class accuracy improvement due to sleep. Each row shows experiments with data reduction for one specific class (shown on the left), with the percentage of reduction shown on the horizontal axis. Each cell shows the class-wise accuracy of the underrepresented class before sleep (top value) and after sleep (bottom value). The color map is based on the change in accuracy, $\Delta=$ After Sleep - Before Sleep. Reds indicate a positive difference (improvement), while blues indicate a negative difference (drop in accuracy). Note, many red squares showing class-wise improvement with only a few blue squares showing class-wise performance loss. }
\label{fig:Imbalanced}
\end{figure}

\subsection{Training with limited and imbalanced data}
We conducted a second set of experiments to test the effect of sleep on training data imbalance, when the overall dataset is limited. This is relevant for many real world scenarios where examples of certain positive or negative classes are not available by virtue of the problem in real world. 
%(e.g. examples of certain types of cancers). 

First, we extracted a subset of images (10\%) from MNIST. Within this limited subset, we explicitly introduced imbalance by reducing the number of images in one selected class, keeping the number of images for all other classes equal and fixed. In this scenario, after the initial training phase with backpropagation, the ANN becomes biased towards classes with more training data at the expense of the class with reduced data. Following this, we tested whether sleep could recover performance in this reduced data class. 

We used both MNIST and FMNIST datasets, while varying the number of images in the training set of one class. The same ANN implementation described in Section \ref{sec:ANN} was used, and the same SRC algorithm was implemented as described in \ref{sec:SRC}.

% \subsection{Training sequential tasks with limited data}

% We conducted a third experiment to investigate continual learning over two different tasks by the ANN in a "low-data" scenario. As before, each time we extracted a small portion (e.g. 1\% - 20\%) of the MNIST/FMNIST databases and split it into two subsets: digits 0-4 (task 1) and digits 5-9 (task 2). 

% The ANN was trained in a sequential fashion: Initially ANN was trained with only digits 0-4 during the first phase over 5 epochs. After completing this first phase, ANN learns to classify digits 0-4 much better than other digits 5-9, which it was not exposed to at all during this initial training. 

% This was followed by a second phase where only task 2 (digits 5-9) was used in training the same ANN over 5 epochs. This second phase of training makes the ANN perform much better in classifying digits 5-9 and essentially "forget" the task 1 (digits 0-4) it learned in first phase.

% These two training phases were followed by the sleep phase as described in Section \ref{sec:SRC}. We analyzed the overall accuracy improvement (across both tasks) due to SRC in this continual learning task. Importantly, our analysis was focused on limited data scenario where only a small subset of training data was available for training.

\subsection{Data Limited Continual Learning }

In our third set of experiments, we utilized the MNIST and FMNIST datasets to examine the influence of SRC on ANN performance in a data limited continual learning paradigm. Here we leveraged 2 tasks, the initial task involved distinguishing classes 0-4 while the subsequent task focused on classes 5-9. The datasets were then randomly sub-sampled such that 1\% to 20\% of the original task data remained and used for model training.

The ANN underwent sequential training with initial training on Task 1 (classes 0-4) followed by training on Task 2 (classes 5-9). Each  MNIST/FMNIST task training phase consisted of 3/5 epochs with a learning rate of 0.06 and a Binary Cross Entropy loss function optimized by Stochastic Gradient Decent (batch size 64). Following initial Task 1 training, the model learned to differentiate classes 0-4 while performance on the untrained Task 2 digits remained poor. After training the second task,  Task 2 performance (classes 5-9) reached a maximum while Task 1 was catastrophically forgotten. These two training phases were followed by SRC (Section \ref{sec:SRC}), after which the ANN was tested again for both tasks. 
%We analyzed the overall improvement across tasks and were able to observe an noticeable improvement in mean task performance Fig. \ref{fig:2T_Heatmap}. This highlighted the benefit of SRC in continual learning context where limited training data is available.

\section{Results}

\subsection{Performance with limited training data}

When the network was trained on the full dataset, the model achieved over 90\% accuracy on both the MNIST and Fashion MNIST datasets. Yet, as illustrated in Figure \ref{fig:fig1}, accuracy significantly drops (blue line) when only a fraction of training data was used. To evaluate the impact of sleep replay on accuracy, Sleep Replay Consolidation (SRC) was employed following initial training with limited data. In brief (for details, see the Methods section), the ANN was trained on limited data then translated to a Spiking Neural Network (SNN) with an identical architecture. This SNN was then stimulated by Poisson-distributed spiking inputs with mean firing rates that matched the mean input element intensity across all datasets used during the preceding training sessions.
%that mirrored the average inputs from the training data. 
During the sleep phase, synaptic weights were adjusted through local Hebbian-type plasticity: strengths were increased if presynaptic activity led to a postsynaptic response, and reduced if postsynaptic activity occurred in the absence of presynaptic activity. Following the sleep phase, the SNN was transformed back into an ANN, and its performance was reassessed either immediately or after an additional fine-tuning on the original limited dataset.

We found a significant improvement in low data range (0.2-10\%) in accuracy for both MNIST and FMNIST datasets after introducing the sleep phase (Figure \ref{fig:fig1}, orange line). We verified these results using 10 and 50 epochs of training. These long training duration experiments revealed that while increasing the amount of training improved baseline performance in data restricted scenarios, sleep was still capable of further improving accuracy. 
%Even when we increased the training duration (number of epochs), the baseline performance before sleep increased, but a noticeable performance gain could be achieved after sleep. 

\begin{figure}[!htbp]
\centering
\includegraphics[width=1.0\columnwidth]{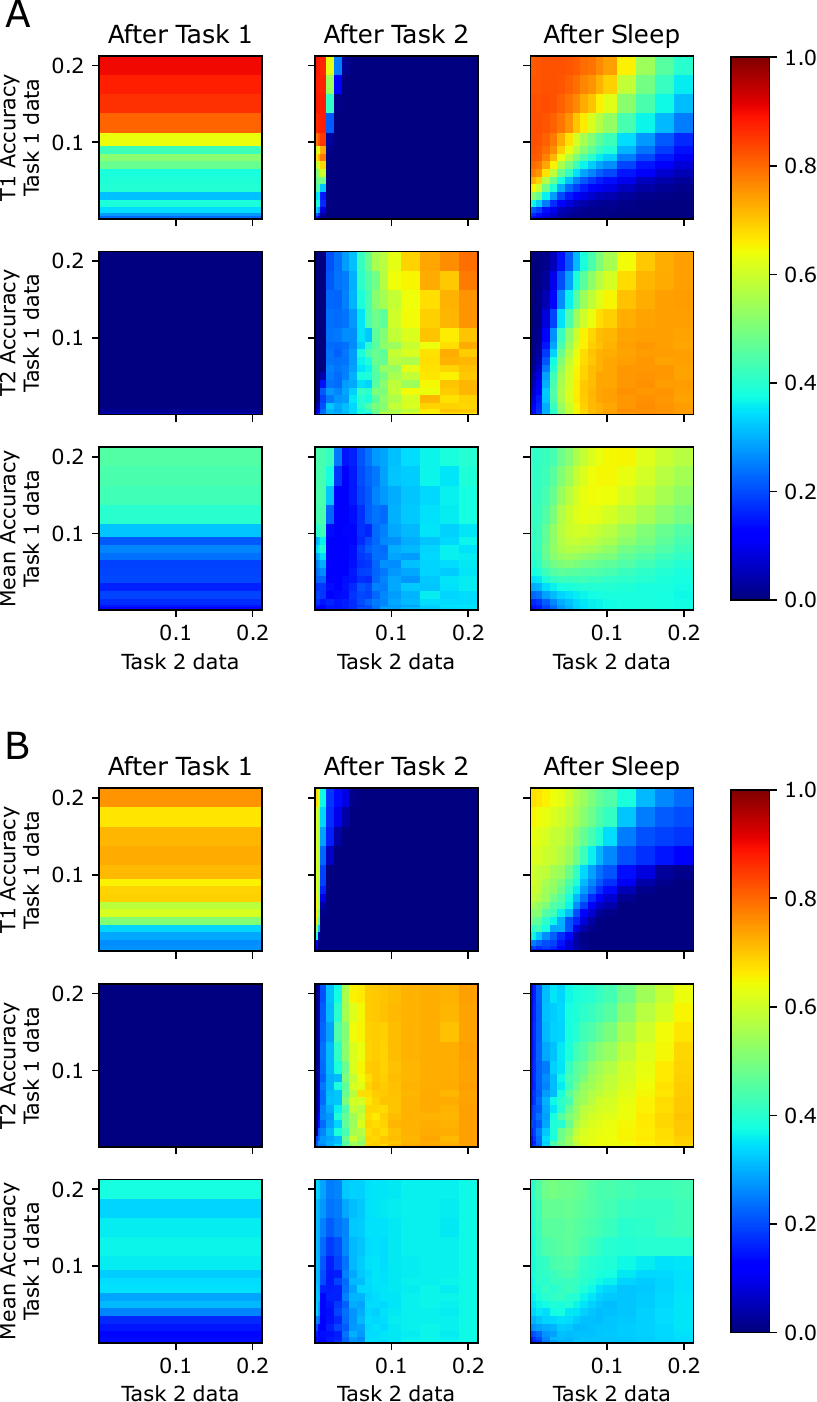} 
\caption{Heatmaps showing accuracy changes after each phase of sequential learning and SRC (columns) on MNIST \textbf{(A)} and FMNIST \textbf{(B)}. In each subplot, X-axis represents amount of T2 training data and Y-axis - amount of T1 training data. The rows show accuracy on T1, T2 and the mean accuracy.}
\label{fig:2T_Heatmap}
\end{figure}

\begin{figure}[!htbp]
\centering
\includegraphics[width=1.0\columnwidth]{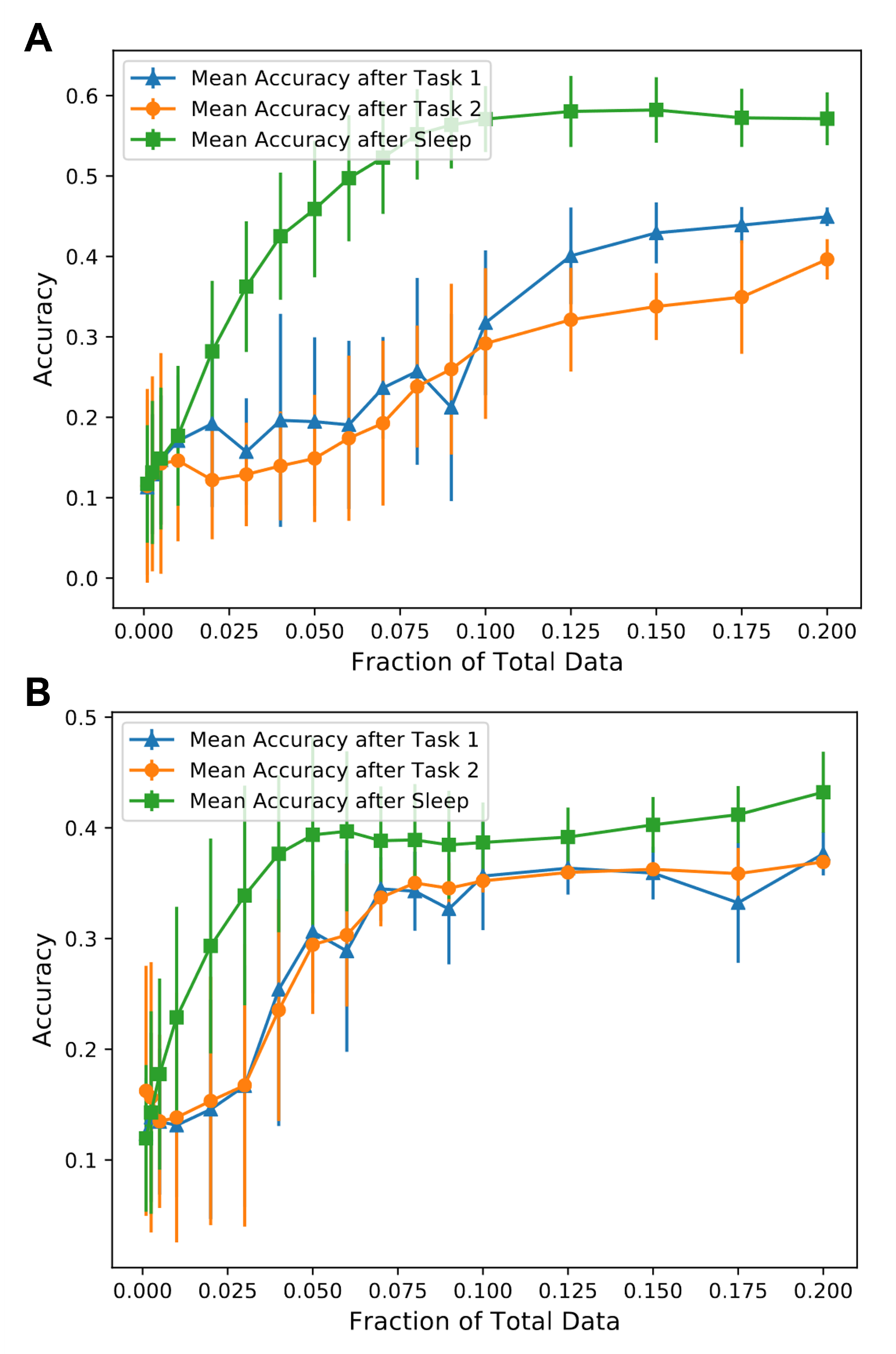} 
\caption{Accuracy for sequential learning tasks on MNIST \textbf{(A)} and FMNIST \textbf{(B)} with mean (lines) and standard deviation (error bars) across 10 trials. X-axis - log of the relative amount of data used for training (e.g., 0.01=1\% of data). Blue - baseline (after ANN training); Orange - baseline + sleep; Green - baseline + sleep + fine-tuning. Note significant gain in overall accuracy after sleep phase on low data.}
\label{fig:2T_Mean}
\end{figure}

Figure \ref{fig:confusion_matrix_MNIST} presents the confusion matrices obtained during the MNIST experiment, prior to and following the sleep phase. For these experiments we employed a 3\% subset of the MNIST dataset for training while using the training parameters detailed in Table \ref{table:nntrain}. Following training, the network demonstrated noteworthy accuracy for only four specific classes, achieving 0.99 for class 0, 0.51 for class 3, 0.71 for class 8, and 0.87 for class 9 (Figure \ref{fig:confusion_matrix_MNIST}A). The remaining classes were predicted inaccurately and misclassified as class 0, 8, or 9. Overall, the prediction accuracy was very low at 0.32. After SRC, there were significant accuracy improvements for most under-performing classes. Specifically there was notable class-wise enhancement for 1, 2, 4, 6 and 7 (Figure \ref{fig:confusion_matrix_MNIST}B) which led to an overall improved accuracy of 0.60.

We obtained similar results for FMNIST. Figure \ref{fig:confusion_matrix_FMNIST} illustrates the confusion matrices generated using a 3\% subset of the FMNIST dataset. Prior to the sleep phase, the overall accuracy stood at 0.38, with only classes 0, 6, 8, and 9 being accurately-classified (Figure \ref{fig:confusion_matrix_FMNIST}A). Following sleep, a notable improvement was observed in the accuracy of other classes (2,3,4,7) (Figure \ref{fig:confusion_matrix_FMNIST}B), culminating in an overall accuracy of 0.59.

These results suggest that not only does SRC improve performance in low data contexts but it proportionally aids under-performing classes yielding a more class balanced model.

While performance improved when there was limited training data, we also observed a slight (10-15\%) decrease in performance when more than approximately 10\% of the data was employed for ANN training.
%As expected, when the full MNIST/FMNIST dataset was used for training, the ANN achieved accuracy of over 90\% in baseline training as shown in the rightmost data points in Figure \ref{fig:fig1} (blue line). When using  larger datasets of higher than 10\% of the total MNIST / FMNIST database, a slight (10-15\%) decrease in performance was observed after sleep phase. 
This performance degradation, however, could be mitigated by introducing a short one epoch fine tuning phase after sleep using the original (limited) training data (Figure \ref{fig:fig1}, green line).

\subsection{Performance with limited and imbalanced data}

Next, we systematically tested the effect of SRC in class imbalanced settings, i.e., in addition to training the network on limited data, one selected class was explicitly underrepresented. 
%When we introduced a class imbalance within a limited training dataset, having a sleep phase helped improve accuracy on the task trained with the reduced number of images. 
Figure \ref{fig:Imbalanced} shows an example of such analysis when we used a 10\%  subset of MNIST training data and further limited the number of images for one selected class during training. In practice, we developed these datasets by first randomly selecting a 10\% subset of the MNIST dataset and ensuring each class had the exact same number of images. Then, for one selected "imbalanced" class, we further restricted the number of instances present causing this class to be underrepresented. 

The fraction of images in the low data class ranged from 10\%  to 100\%  of the original training data subset (i.e. 1\% to 10\% of the class from the original full MNIST).

We found that class-wise model performance was more robust to data reduction for some classes when compared to others. For example, digit 0 showed high class-wise accuracy when more than 70-80\% digit 0's were used for training (i.e., more than 7-8\% of digit 0's in the total dataset), while digit 5 had very low performance even when all data were used (i.e., 10\% of 5's in the total dataset). After SRC, most classes (except digit 8) showed a positive improvement in class-wise accuracy (Figure \ref{fig:Imbalanced}). 
%when the amount of training data used for the class is 40-90\%  of the overall 10\%  subset of the training data used for the other classes. In Figure \ref{fig:Imbalanced}, many red squares showing class-wise improvement with only a few blue squares showing class-wise performance loss. 
The magnitude of the gain and the range of data reduction where the  gain was observed were varied between digits likely because of the different sensitivity to data reduction. Similar results (not shown) were observed with the FMNIST dataset.

\subsection{Performance during sequential learning tasks with limited data}

We next tested how sleep replay can affect performance of models trained on sequential data limited tasks. A previous study \cite{Tadros_NC2022} reported that SRC can protect older tasks damaged by the recent training, however, it remains unknown if SRC can accomplish the same effect when training data are limited. It is also unknown if the same SRC hyper-parameters can both protect old tasks' performance while simultaneously improving recent under-trained task performance.

%ANNs have been shown to suffer from catastrophic forgetting for various standard image datasets including MNIST and Fashion MNIST \cite{kemker2018measuring}. 
To test SRC in a sequential learning paradigm, we created 2 tasks (per dataset) for the MNIST and Fashion MNIST where the first task consisted of the first 5 classes and the second task comprised of the last 5. These two tasks (T1 and T2)  were trained sequentially followed by a sleep phase. The amount of data used to train each task was varied independently in the range 0-20\% of the full datasets.

Figure \ref{fig:2T_Heatmap} shows results for MNIST and Fashion MNIST datasets. After T1 training, the network increased accuracy for that task (top left plots in Fig. \ref{fig:2T_Heatmap}A,B) but had zero performance on T2 that was not yet trained (middle left plots in Fig. \ref{fig:2T_Heatmap}A,B). Subsequent T2 training led to increase of T2 performance (middle middle plots in Fig. \ref{fig:2T_Heatmap}A,B) but T1 was suffered from catastrophic forgetting and T1 performance reduced to zero, except when T2 training data were very low (top middle plots in Fig. \ref{fig:2T_Heatmap}A,B). Finally, SRC implemented after T2 training rescued T1, except when T2 data significantly exceeded T1 data (top right plots in Fig. \ref{fig:2T_Heatmap}A,B). T2 performance was further increased for MNIST (middle right plot in Fig. \ref{fig:2T_Heatmap}A) and slightly decreased for FMNIST (middle right plot in Fig. \ref{fig:2T_Heatmap}B).

\begin{figure}[t] 
    \centering
  % \subfloat{
  %       \label{fig:fig_a}
  %       \includegraphics[width=0.45\linewidth]{Figures/mnist/allSumSpikes.pdf} 
  %      }
  % \subfloat{
  %       \label{fig:fig_b}
  %       \includegraphics[width=0.45\linewidth]{{Figures/fmnist/allSumSpikes.pdf}        
  %       }    
   \includegraphics[width=1.0\linewidth]{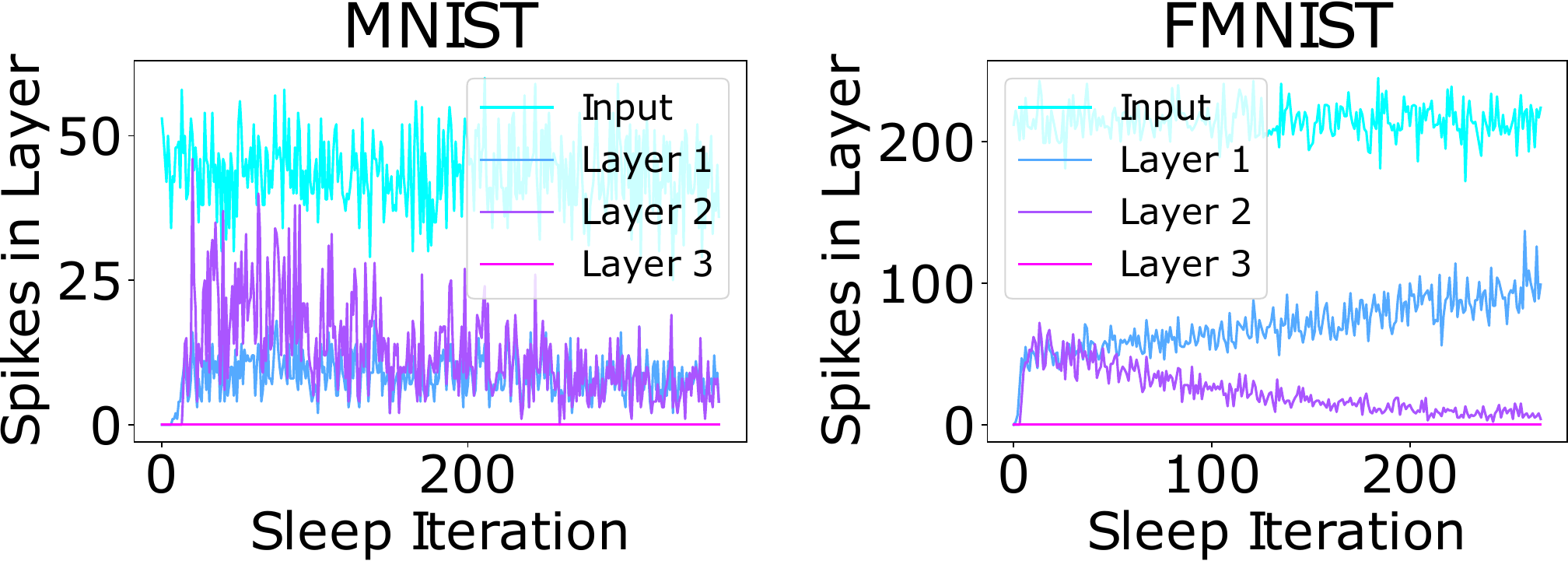}
  \caption{Layer-wise activity during SRC for MNIST \textbf{(left)} and FMNIST \textbf{(right)}. It can be seen that all layers are active except the output layer in both the MNIST and FMNIST networks implying the benefits of SRC are due to enhancing the networks ability to extract relevant features.}
  \label{fig:srcActivity} 
\end{figure}

Figure \ref{fig:2T_Mean} presents the same results but for the specific case of equal amounts of training data for T1 and T2 (i.e., diagonal in Figure \ref{fig:2T_Heatmap}) with the Y-axis denoting mean accuracy between two tasks and X-axis denoting amount of data. Overall, we observed an increase in the mean accuracy in the range 15-30\% for MNIST and 5-15\% for FMNIST with the exact improvement dependent on the amount of training data.

\subsection{Network dynamics during SRC}

In an attempt to gain further insight as to why SRC was capable of improving  under-trained model performance, we analyzed the sleep stage for the single task scenario. We first measured network activity by examining the instantaneous layer-wise firing rates. It can be observed that all layers, excluding the output layer, in both the MNIST and FMNIST networks were highly active (Figure \ref{fig:srcActivity}). The lack of output layer activity along with the previously described Hebbian learning rules means the output layer received no modifications during the entirety of SRC, all benefits were therefore due to hidden layer modifications. 

\begin{figure} 
    \centering
  % \subfloat{
  %       \label{fig:srcspikeRasterMNIST}
  %       \includegraphics[width=0.45\linewidth]{Figures/mnist/spikes-mnist.pdf}       
  %      }
  % \subfloat{
  %       \label{fig:srcspikeRasterFMNIST}
  %       \includegraphics[width=0.45\linewidth]{Figures/fmnist/spikes-fmnist.pdf}        
  %       }  
        \includegraphics[width=1.0\linewidth]{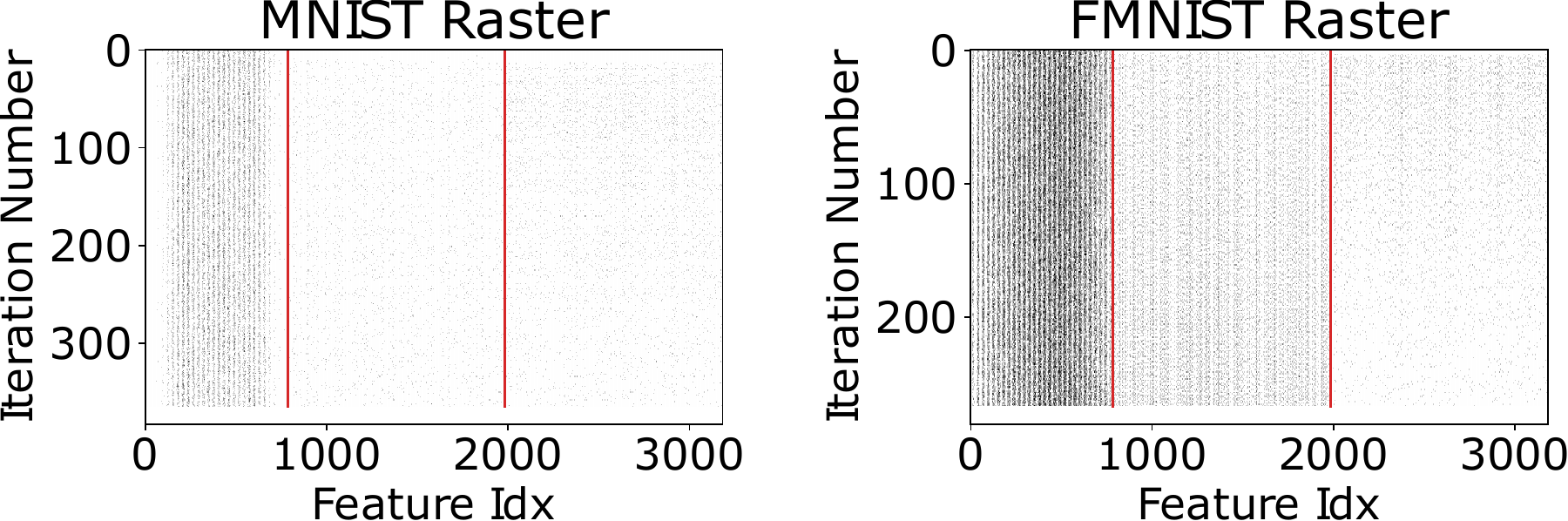}        
        
  \caption{MNIST \textbf{(left)} and FMNIST \textbf{(right)} spike rasters for input and hidden layers (denoted by red lines). While generated input remains at a consistent level (left of first red lines), hidden layer activity begins high and slightly decays over the course of sleep (right of first red lines)}
  \label{fig:srcspikeRaster} 
\end{figure}

Analysis of individual hidden layer neuron activity revealed a progressive decay of activity over the course of the sleep phase (Figure \ref{fig:srcspikeRaster}). This was more obvious in the second hidden layer. Thus, 
%in agreement with previous report \cite{Tadros_NC2022} for multi-task case with full amount of training data, 
task performance improved by SRC reducing the overall level of activity in the network thereby forming sparser, but presumably more contrasting, representations between tasks. 

\begin{figure}[b] 
    \centering
  % \subfloat{
  %       \label{fig:rasterMnist}
  %       \includegraphics[width=0.45\linewidth]{Figures/mnist/weightDiffHistNetwork.pdf}       
  %      }
  % \subfloat{
  %       \label{fig:rasterFMnist}
  %       \includegraphics[width=0.45\linewidth]{Figures/fmnist/weightDiffHistNetwork.pdf}        
  %       }  
        \includegraphics[width=1.0\linewidth]{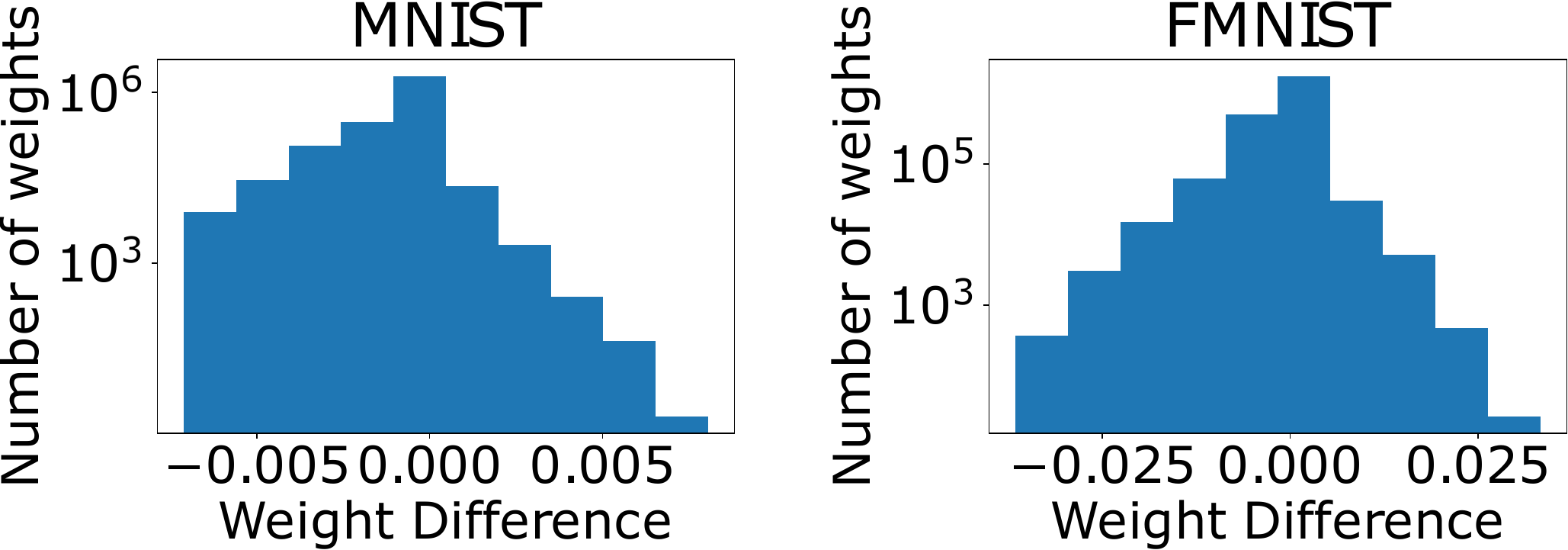}       
        
  \caption{ Weight difference distributions for MNIST \textbf{(left)} and FMNIST \textbf{(right)}. Plots highlight that many synapses decrease in strength implying performance benefits are a result of suppressing incorrect signals.}
  \label{fig:srcWeightDiffDist} 
\end{figure}

To confirm this prediction, we next examined the magnitude of weight modifications. Figure \ref{fig:srcWeightDiffDist} shows a histogram of the cumulative weight perturbation each synapse received during sleep. 
%Overall, we observed increase in some weights and decrease on others, with some overall bias towards reduction of synaptic weights.  
Although a small number of critical synapses showed an increase in strength (positive weight deltas) most synapses decreased their sensitivity (Figure \ref{fig:srcWeightDiffDist} shows large number of synapses with negative weight deltas; note log scale in Y-axis).

\section{Discussion}

% I moved section on the weights analysis from Discussion to the end of Results. Weights analysis is still a result and Discussion section should be fully dedicated to discussing the results and not introducing new results
Although artificial neural networks (ANNs) have recently begun to approach human performance in various tasks, from intricate games \cite{silver2018general} to image classification \cite{krizhevsky2012imagenet} to language processing \cite{Liu2023SummaryOC}, they exhibit several limitations. These include catastrophic forgetting, lack of generalization, and dependency on substantial amounts of training data. Catastrophic forgetting \cite{mcclelland1995there,french1999catastrophic,mccloskey1989catastrophic} is the phenomenon where a system cannot learn new information without losing previously acquired knowledge. Consequently, even if new data that could enhance a model's performance are available, direct training on these new datasets often results in damage to previously learned information. Additionally, ANNs may develop bias when training data is limited and imbalanced, leading to models that are unfairly prejudiced against certain categories. This poses risks in sensitive applications like healthcare, where it could result in misdiagnoses, or in legal settings, where it might contribute to unequal treatment \cite{Norori2021AddressingBias, PS2023100165}. Addressing these biases is crucial for creating fair and reliable AI systems.

Here, we have tested an unsupervised sleep replay consolidation (SRC) algorithm on artificial neural networks trained incrementally with limited and imbalanced datasets. We found that SRC effectively mitigates catastrophic forgetting and improves accuracy when only a very small fraction of data is used for training or when data is class-imbalanced. Effectively, SRC was able to shift the transition point from "low" to "high" accuracy as the amount of training data increases. The relationship between training data and accuracy has been widely studied (e.g., \cite{paul2021deep}), and our study provides evidence that the transition to high accuracy during learning could be influenced by spontaneously generated unsupervised replay. One advantage of SRC is that it is complementary to fine-tuning and other gradient-based methods and thus can be used in conjunction with these methods. 

%SRC emulates essential features of biological sleep replay, which include: (a) Neurons exhibiting spontaneous activity during sleep, resulting in the replay of previously learned activation patterns. (b) Unsupervised Hebbian-type plasticity, which modifies the connectivity matrix to enhance task-specific connections and prune excessive connections between neurons. This process enhances the sparsity of representations and reduces overlapping representations among different input classes.

Existing methods to prevent catastrophic forgetting generally fall into two categories: rehearsal and regularization techniques \cite{kemker2018measuring}. Rehearsal methods involve incorporating previously learned data, either stored or generated, alongside new inputs during subsequent training sessions to mitigate forgetting \cite{hayes2019memory, robins1995catastrophic, van2020brain, shin2017continual, buzzega2020dark, buzzega2021rethinking}. While effective, as the number of previously learned tasks increases, this approach may necessitate increasingly complex generative networks capable of reproducing all previously acquired knowledge.

Regularization methods aim to modify plasticity rules, maintaining important weights from previous tasks by adding constraints to gradient descent \cite{li2017learning}. Notably, the Elastic Weight Consolidation (EWC), Synaptic Intelligence (SI) and Orthogonal Weight Modification (OWM)  techniques penalize weight updates for prior tasks \cite{kirkpatrick2017overcoming, zenke2017continual,zeng2019continual}. While these methods show promise in preventing catastrophic forgetting across various tasks, such as MNIST permutations, Split MNIST, and Atari games, they may not perform optimally in class-incremental learning scenarios, where one input class is learned at a time \cite{van2020brain}. OWM has demonstrated greater success than EWC and SI in class-incremental learning tasks.

Methods to improve network accuracy when training data are limited include data augmentation \cite{shorten2019survey}, pre-training on other datasets \cite{zhuang2020comprehensive} or alternative architectures such as neural tangent kernel \cite{arora2019harnessing}. It is worth noting, that these methods should be applied independently from those developed to enable continual learning leading to complex models with many hyperparameters to be tuned.

SRC has been proposed as an unsupervised method applicable to solving various tasks, including catastrophic forgetting \cite{Tadros_NC2022}, lack of generalization \cite{Delanois_ICMLA23}, and low accuracy with limited training data \cite{Bazhenov_AAAI24}. However, the advantages of SRC were reported in independent studies, each using different sets of hyperparameters tuned to optimize performance for a specific task. Here, we report that SRC can accomplish several independent tasks together in the same model and using the same set of hyperparameters. These results are consistent with the known role of sleep in memory and learning.

Sleep has been shown to play a critical role in memory consolidation, generalization, and the transfer of knowledge in biological systems \cite{ji2007coordinated,walker2004sleep,lewis2011overlapping}. During  sleep, there is a reactivation of neurons involved in previously learned activity \cite{stickgold2005sleep}. Because sleep reactivation depends on the previously learned synaptic connectivity matrix, it invokes the same spatio-temporal patterns of neuronal firing as the patterns observed during preceding training in the awake state \cite{wilson1994reactivation}.  Thus, sleep reactivation, or replay, would strengthen synapses involved in a previously learned task. It was reported that plasticity changes during sleep can increase a brain’s ability to reduce memory interference \cite{gonzalez2020can}, to form connections between memories and to generalize knowledge learned during the awake state \cite{payne2009role}, to link semantic categories based on a single episodic event to enable one-shot learning \cite{witkowski2020examining}.

Analysis of synaptic weight dynamics revealed that while SRC increased the strength for a subset of presumably critical synapses, many others were weakened. This suggests that the overall increase in accuracy after SRC was a result of sparser responses. Thus, SRC may be improving feature representations by causing hidden layer neurons to be less sensitive to certain stimuli. Intuitively, in a data-limited context, the training distribution may have many outliers among the small number of samples, thereby causing the model to pick up irrelevant features only present in a small percentage of examples. SRC then selectively suppresses the strength of many synapses while maintaining a few, thereby reducing irrelevant feature sensitivity and leading the model to focus on the most common and therefore predictive features. In multiple task scenarios, SRC may reduce weights biased towards the most recent tasks, thus allowing older tasks to be correctly classified as long as information about these tasks is still present in the synaptic weights matrix.

In sum, our study sheds light on a potential synaptic weight dynamics strategy employed by the brain during sleep to enhance memory performance for continual learning when training data are limited or imbalanced. Applied to ANNs, sleep-like replay improves performance in a completely unsupervised manner, requiring no additional data, and can be applied to already trained models.

\section*{Acknowledgment}

\bibliographystyle{IEEEtran}
\bibliography{references,references-NC}

\begin{thebibliography}{10}
\providecommand{\url}[1]{#1}
\csname url@samestyle\endcsname
\providecommand{\newblock}{\relax}
\providecommand{\bibinfo}[2]{#2}
\providecommand{\BIBentrySTDinterwordspacing}{\spaceskip=0pt\relax}
\providecommand{\BIBentryALTinterwordstretchfactor}{4}
\providecommand{\BIBentryALTinterwordspacing}{\spaceskip=\fontdimen2\font plus
\BIBentryALTinterwordstretchfactor\fontdimen3\font minus \fontdimen4\font\relax}
\providecommand{\BIBforeignlanguage}[2]{{%
\expandafter\ifx\csname l@#1\endcsname\relax
\typeout{** WARNING: IEEEtran.bst: No hyphenation pattern has been}%
\typeout{** loaded for the language `#1'. Using the pattern for}%
\typeout{** the default language instead.}%
\else
\language=\csname l@#1\endcsname
\fi
#2}}
\providecommand{\BIBdecl}{\relax}
\BIBdecl

\bibitem{RN4279}
\BIBentryALTinterwordspacing
D.~Kudithipudi, M.~Aguilar-Simon, J.~Babb, M.~Bazhenov, D.~Blackiston, J.~Bongard, A.~P. Brna, S.~C. Raja, N.~Cheney, J.~Clune, A.~Daram, S.~Fusi, P.~Helfer, L.~Kay, N.~Ketz, Z.~Kira, S.~Kolouri, J.~L. Krichmar, S.~Kriegman, M.~Levin, S.~Madireddy, S.~Manicka, A.~Marjaninejad, B.~McNaughton, R.~Miikkulainen, Z.~Navratilova, T.~Pandit, A.~Parker, P.~K. Pilly, S.~Risi, T.~J. Sejnowski, A.~Soltoggio, N.~Soures, A.~S. Tolias, D.~Urbina-Melendez, F.~J. Valero-Cuevas, G.~M. van~de Ven, J.~T. Vogelstein, F.~Wang, R.~Weiss, A.~Yanguas-Gil, X.~Y. Zou, and H.~Siegelmann, ``Biological underpinnings for lifelong learning machines,'' \emph{Nature Machine Intelligence}, vol.~4, no.~3, pp. 196--210, 2022. [Online]. Available: \url{<Go to ISI>://WOS:000772442700002}
\BIBentrySTDinterwordspacing

\bibitem{hayes2021replay}
T.~L. Hayes, G.~P. Krishnan, M.~Bazhenov, H.~T. Siegelmann, T.~J. Sejnowski, and C.~Kanan, ``Replay in deep learning: Current approaches and missing biological elements,'' \emph{Neural Computation}, vol.~33, no.~11, pp. 2908--2950, 2021.

\bibitem{shorten2019survey}
C.~Shorten and T.~M. Khoshgoftaar, ``A survey on image data augmentation for deep learning,'' \emph{Journal of big data}, vol.~6, no.~1, pp. 1--48, 2019.

\bibitem{zhuang2020comprehensive}
F.~Zhuang, Z.~Qi, K.~Duan, D.~Xi, Y.~Zhu, H.~Zhu, H.~Xiong, and Q.~He, ``A comprehensive survey on transfer learning,'' \emph{Proceedings of the IEEE}, vol. 109, no.~1, pp. 43--76, 2020.

\bibitem{arora2019harnessing}
S.~Arora, S.~S. Du, Z.~Li, R.~Salakhutdinov, R.~Wang, and D.~Yu, ``Harnessing the power of infinitely wide deep nets on small-data tasks,'' \emph{arXiv preprint arXiv:1910.01663}, 2019.

\bibitem{kemker2018measuring}
R.~Kemker, M.~McClure, A.~Abitino, T.~L. Hayes, and C.~Kanan, ``Measuring catastrophic forgetting in neural networks,'' in \emph{Thirty-second AAAI conference on artificial intelligence}, 2018.

\bibitem{hayes2019memory}
T.~L. Hayes, N.~D. Cahill, and C.~Kanan, ``Memory efficient experience replay for streaming learning,'' in \emph{2019 International Conference on Robotics and Automation (ICRA)}.\hskip 1em plus 0.5em minus 0.4em\relax IEEE, 2019, pp. 9769--9776.

\bibitem{robins1995catastrophic}
A.~Robins, ``Catastrophic forgetting, rehearsal and pseudorehearsal,'' \emph{Connection Science}, vol.~7, no.~2, pp. 123--146, 1995.

\bibitem{van2020brain}
G.~M. van~de Ven, H.~T. Siegelmann, and A.~S. Tolias, ``Brain-inspired replay for continual learning with artificial neural networks,'' \emph{Nature communications}, vol.~11, no.~1, pp. 1--14, 2020.

\bibitem{shin2017continual}
H.~Shin, J.~K. Lee, J.~Kim, and J.~Kim, ``Continual learning with deep generative replay,'' in \emph{Advances in Neural Information Processing Systems}, 2017, pp. 2990--2999.

\bibitem{buzzega2020dark}
P.~Buzzega, M.~Boschini, A.~Porrello, D.~Abati, and S.~Calderara, ``Dark experience for general continual learning: a strong, simple baseline,'' \emph{Advances in neural information processing systems}, vol.~33, pp. 15\,920--15\,930, 2020.

\bibitem{buzzega2021rethinking}
P.~Buzzega, M.~Boschini, A.~Porrello, and S.~Calderara, ``Rethinking experience replay: a bag of tricks for continual learning,'' in \emph{2020 25th International Conference on Pattern Recognition (ICPR)}.\hskip 1em plus 0.5em minus 0.4em\relax IEEE, 2021, pp. 2180--2187.

\bibitem{li2017learning}
Z.~Li and D.~Hoiem, ``Learning without forgetting,'' \emph{IEEE transactions on pattern analysis and machine intelligence}, vol.~40, no.~12, pp. 2935--2947, 2017.

\bibitem{kirkpatrick2017overcoming}
J.~Kirkpatrick, R.~Pascanu, N.~Rabinowitz, J.~Veness, G.~Desjardins, A.~A. Rusu, K.~Milan, J.~Quan, T.~Ramalho, A.~Grabska-Barwinska \emph{et~al.}, ``Overcoming catastrophic forgetting in neural networks,'' \emph{Proceedings of the national academy of sciences}, vol. 114, no.~13, pp. 3521--3526, 2017.

\bibitem{stickgold2005sleep}
R.~Stickgold, ``Sleep-dependent memory consolidation,'' \emph{Nature}, vol. 437, no. 7063, pp. 1272--1278, 2005.

\bibitem{lewis2011overlapping}
P.~A. Lewis and S.~J. Durrant, ``Overlapping memory replay during sleep builds cognitive schemata,'' \emph{Trends in cognitive sciences}, vol.~15, no.~8, pp. 343--351, 2011.

\bibitem{steriade1993thalamocortical}
M.~Steriade, D.~A. McCormick, and T.~J. Sejnowski, ``Thalamocortical oscillations in the sleeping and aroused brain,'' \emph{Science}, vol. 262, no. 5134, pp. 679--685, 1993.

\bibitem{krishnan2016cellular}
G.~P. Krishnan, S.~Chauvette, I.~Shamie, S.~Soltani, I.~Timofeev, S.~S. Cash, E.~Halgren, and M.~Bazhenov, ``Cellular and neurochemical basis of sleep stages in the thalamocortical network,'' \emph{Elife}, vol.~5, p. e18607, 2016.

\bibitem{wilson1994reactivation}
M.~A. Wilson and B.~L. McNaughton, ``Reactivation of hippocampal ensemble memories during sleep,'' \emph{Science}, vol. 265, no. 5172, pp. 676--679, 1994.

\bibitem{wei2016synaptic}
Y.~Wei, G.~P. Krishnan, and M.~Bazhenov, ``Synaptic mechanisms of memory consolidation during sleep slow oscillations,'' \emph{Journal of Neuroscience}, vol.~36, no.~15, pp. 4231--4247, 2016.

\bibitem{lewis2018memory}
P.~A. Lewis, G.~Knoblich, and G.~Poe, ``How memory replay in sleep boosts creative problem-solving,'' \emph{Trends in cognitive sciences}, vol.~22, no.~6, pp. 491--503, 2018.

\bibitem{hennevin1995processing}
E.~Hennevin, B.~Hars, C.~Maho, and V.~Bloch, ``Processing of learned information in paradoxical sleep: relevance for memory,'' \emph{Behavioural brain research}, vol.~69, no. 1-2, pp. 125--135, 1995.

\bibitem{rasch2013sleep}
B.~Rasch and J.~Born, ``About sleep's role in memory,'' \emph{Physiological reviews}, 2013.

\bibitem{mednick2011opportunistic}
S.~C. Mednick, D.~J. Cai, T.~Shuman, S.~Anagnostaras, and J.~T. Wixted, ``An opportunistic theory of cellular and systems consolidation,'' \emph{Trends in neurosciences}, vol.~34, no.~10, pp. 504--514, 2011.

\bibitem{paller2004memory}
K.~A. Paller and J.~L. Voss, ``Memory reactivation and consolidation during sleep,'' \emph{Learning \& Memory}, vol.~11, no.~6, pp. 664--670, 2004.

\bibitem{oudiette2013role}
D.~Oudiette, J.~W. Antony, J.~D. Creery, and K.~A. Paller, ``The role of memory reactivation during wakefulness and sleep in determining which memories endure,'' \emph{Journal of Neuroscience}, vol.~33, no.~15, pp. 6672--6678, 2013.

\bibitem{gonzalez2020can}
O.~C. Gonz{\'a}lez, Y.~Sokolov, G.~P. Krishnan, J.~E. Delanois, and M.~Bazhenov, ``Can sleep protect memories from catastrophic forgetting?'' \emph{Elife}, vol.~9, p. e51005, 2020.

\bibitem{golden2020sleep}
R.~Golden, J.~E. Delanois, P.~Sanda, and M.~Bazhenov, ``Sleep prevents catastrophic forgetting in spiking neural networks by forming joint synaptic weight representations,'' \emph{PLoS Computational Biology}, vol.~18, no.~11, p. e1010628, 2022.

\bibitem{Tadros_NC2022}
T.~Tadros, G.~Krishnan, R.~Ramyaa, and M.~Bazhenov, ``Sleep-like unsupervised replay reduces catastrophic forgetting in artificial neural networks,'' \emph{Nature Communications}, vol.~13, no.~1, p. 7742, 2022.

\bibitem{diehl2015fast}
P.~U. Diehl, D.~Neil, J.~Binas, M.~Cook, S.-C. Liu, and M.~Pfeiffer, ``Fast-classifying, high-accuracy spiking deep networks through weight and threshold balancing,'' in \emph{2015 International Joint Conference on Neural Networks (IJCNN)}.\hskip 1em plus 0.5em minus 0.4em\relax ieee, 2015, pp. 1--8.

\bibitem{lecun1998gradient}
Y.~LeCun, L.~Bottou, Y.~Bengio, and P.~Haffner, ``Gradient-based learning applied to document recognition,'' \emph{Proceedings of the IEEE}, vol.~86, no.~11, pp. 2278--2324, 1998.

\bibitem{xiao2017fashion}
H.~Xiao, K.~Rasul, and R.~Vollgraf, ``Fashion-mnist: a novel image dataset for benchmarking machine learning algorithms,'' \emph{arXiv preprint arXiv:1708.07747}, 2017.

\bibitem{silver2018general}
D.~Silver, T.~Hubert, J.~Schrittwieser, I.~Antonoglou, M.~Lai, A.~Guez, M.~Lanctot, L.~Sifre, D.~Kumaran, T.~Graepel \emph{et~al.}, ``A general reinforcement learning algorithm that masters chess, shogi, and go through self-play,'' \emph{Science}, vol. 362, no. 6419, pp. 1140--1144, 2018.

\bibitem{krizhevsky2012imagenet}
A.~Krizhevsky, I.~Sutskever, and G.~Hinton, ``Imagenet classification with deep convolutional neural networks,'' in \emph{Advances in neural information processing systems}, 2012.

\bibitem{Liu2023SummaryOC}
\BIBentryALTinterwordspacing
Y.~Liu, T.~Han, S.~Ma, J.~Zhang, Y.~Yang, J.~Tian, H.~He, A.~Li, M.~He, Z.~Liu, Z.~Wu, L.~Zhao, D.~Zhu, X.~Li, N.~Qiang, D.~Shen, T.~Liu, and B.~Ge, ``Summary of chatgpt-related research and perspective towards the future of large language models,'' \emph{Meta-Radiology}, vol. 100017, p.~21, 2023, arXiv:2304.01852v4 [cs.CL]. [Online]. Available: \url{https://doi.org/10.48550/arXiv.2304.01852}
\BIBentrySTDinterwordspacing

\bibitem{mcclelland1995there}
J.~L. McClelland, B.~L. McNaughton, and R.~C. O'Reilly, ``Why there are complementary learning systems in the hippocampus and neocortex: insights from the successes and failures of connectionist models of learning and memory.'' \emph{Psychological review}, vol. 102, no.~3, p. 419, 1995.

\bibitem{french1999catastrophic}
R.~M. French, ``Catastrophic forgetting in connectionist networks,'' \emph{Trends in cognitive sciences}, vol.~3, no.~4, pp. 128--135, 1999.

\bibitem{mccloskey1989catastrophic}
M.~McCloskey and N.~J. Cohen, ``Catastrophic interference in connectionist networks: The sequential learning problem,'' in \emph{Psychology of learning and motivation}.\hskip 1em plus 0.5em minus 0.4em\relax Elsevier, 1989, vol.~24, pp. 109--165.

\bibitem{Norori2021AddressingBias}
N.~Norori, Q.~Hu, F.~M. Aellen, F.~D. Faraci, and A.~Tzovara, ``Addressing bias in big data and ai for health care: A call for open science,'' \emph{Patterns}, vol.~2, no.~10, p. 100347, 2021.

\bibitem{PS2023100165}
P.~Varsha, ``How can we manage biases in artificial intelligence systems--a systematic literature review,'' \emph{International Journal of Information Management Data Insights}, vol.~3, no.~1, p. 100165, 2023.

\bibitem{paul2021deep}
M.~Paul, S.~Ganguli, and G.~K. Dziugaite, ``Deep learning on a data diet: Finding important examples early in training,'' \emph{Advances in Neural Information Processing Systems}, vol.~34, pp. 20\,596--20\,607, 2021.

\bibitem{zenke2017continual}
F.~Zenke, B.~Poole, and S.~Ganguli, ``Continual learning through synaptic intelligence,'' in \emph{Proceedings of the 34th International Conference on Machine Learning-Volume 70}.\hskip 1em plus 0.5em minus 0.4em\relax JMLR. org, 2017, pp. 3987--3995.

\bibitem{zeng2019continual}
G.~Zeng, Y.~Chen, B.~Cui, and S.~Yu, ``Continual learning of context-dependent processing in neural networks,'' \emph{Nature Machine Intelligence}, vol.~1, no.~8, pp. 364--372, 2019.

\bibitem{Delanois_ICMLA23}
J.~E. Delanois, A.~Ahuja, G.~Krishnan, T.~Tadros, and M.~Bazhenov, ``Improving robustness of convolutional networks through sleep-like replay,'' in \emph{2023 22nd IEEE International Conference on Machine Learning and Applications (ICMLA)}.\hskip 1em plus 0.5em minus 0.4em\relax IEEE, 2023.

\bibitem{Bazhenov_AAAI24}
A.~Bazhenov, P.~Dewasurendra, G.~Krishnan, and J.~E. Delanois, ``Sleep-like unsupervised replay improves performance when data are limited or unbalanced (student abstract),'' in \emph{Proceedings of the AAAI Conference on Artificial Intelligence}, 2024 in press.

\bibitem{ji2007coordinated}
D.~Ji and M.~A. Wilson, ``Coordinated memory replay in the visual cortex and hippocampus during sleep,'' \emph{Nature neuroscience}, vol.~10, no.~1, pp. 100--107, 2007.

\bibitem{walker2004sleep}
M.~P. Walker and R.~Stickgold, ``Sleep-dependent learning and memory consolidation,'' \emph{Neuron}, vol.~44, no.~1, pp. 121--133, 2004.

\bibitem{payne2009role}
J.~D. Payne and et~al., ``The role of sleep in false memory formation,'' \emph{Neurobiology of Learning and Memory}, vol.~92, no.~3, pp. 327--334, 2009.

\bibitem{witkowski2020examining}
S.~Witkowski, E.~Schechtman, and K.~A. Paller, ``Examining sleep’s role in memory generalization and specificity through the lens of targeted memory reactivation,'' \emph{Current Opinion in Behavioral Sciences}, vol.~33, pp. 86--91, 2020.

\end{thebibliography}
%\bibliography{references-NC}

\end{document}